\documentclass[a4paper, conference, onecolumn]{IEEEtran}
\IEEEoverridecommandlockouts
\usepackage{cite}
\usepackage{amsmath,amssymb,amsfonts}
\usepackage{algorithmic}
\usepackage{graphicx}
\usepackage{textcomp}
\usepackage{xcolor}
\usepackage{booktabs}
\usepackage{multirow}
\usepackage[normalem]{ulem}
\usepackage{enumitem}
\usepackage{newtxmath}
\usepackage[a4paper,top=2.1cm,bottom=2.98cm,left=2cm,right=2cm]{geometry}

\def\BibTeX{{\rm B\kern-.05em{\sc i\kern-.025em b}\kern-.08em
    T\kern-.1667em\lower.7ex\hbox{E}\kern-.125emX}}

\begin{document}

\title{Learning Dynamic Graphs via Tensorized and Lightweight Graph Convolutional Networks
}

\author{

\IEEEauthorblockN{Minglian Han}
\IEEEauthorblockA{
\textit{College of Computer and Information Science, Southwest University, Chongqing, China} \\
hanmlian@email.swu.edu.cn}
}


\maketitle

\begin{abstract}

A dynamic graph (DG) is frequently encountered in numerous real-world scenarios. Consequently, A dynamic graph convolutional network (DGCN) has been successfully applied to perform precise representation learning on a DG. However, conventional DGCNs typically consist of a static GCN coupled with a sequence neural network (SNN) to model spatial and temporal patterns separately. This decoupled modeling mechanism inherently disrupts the intricate spatio-temporal dependencies. To address the issue, this study proposes a novel Tensorized Lightweight Graph Convolutional Network (TLGCN) for accurate dynamic graph learning. It mainly contains the following two key concepts: a) designing a novel spatio-temporal information propagation method for joint propagation of spatio-temporal information based on the tensor M-product framework; b) proposing a tensorized lightweight graph convolutional network based on the above method, which significantly reduces the memory occupation of the model by omitting complex feature transformation and nonlinear activation. Numerical experiments on four real-world datasets demonstrate that the proposed TLGCN outperforms the state-of-the-art models in the weight estimation task on DGs.

\end{abstract}


\section{Introduction}
Dynamic graphs (DGs) are widely involved in real-world complex systems \cite{hu2023fcan, jin2022neural, yuan2020temporal} like social networks \cite{b1, yuan2024fuzzy}, transportation networks  \cite{b2}, financial transaction networks  \cite{b3, luo2021novel}, and biological networks  \cite{b4, luo2023predicting}. Compared with the static graphs  \cite{b5, b7, b8}, the interactions of nodes in DGs usually change over time, and the features of nodes may also change dynamically in a specific time period to form certain temporal relational features  \cite{b11, yuan2022kalman}, i.e., the change in traffic flow  \cite{b12, lin20243d, xu2023hrst} or the evolution of social network user behavior  \cite{b13, yuan2024adaptive}. From this perspective, a DG is not a straightforward extension of a static graph \cite{liu2023symmetry} since it introduces temporal information based on static topology  \cite{b14}. Hence, achieving accurate learning of DGs has become a hot yet arduous issue.

Dynamic graph convolutional networks (DGNNs)  \cite{b15} have attracted much attention in recent years due to their excellent ability in DG learning. Therefore, DGNNs are commonly utilized in downstream tasks in real scenarios such as financial fraud detection \cite{b3, zhou2023cryptocurrency}, traffic flow prediction  \cite{b12, yang2024latent}, and real-time social network analysis  \cite{b13, yuan2023adaptive, yuan2020generalized, chen2021hyper}. Note that DGNN mainly typically employs sequential neural networks (e.g., Transformer \cite{b21} or RNN \cite{b22, yan2023modified}) to explore temporal dependency, and adopt a static graph convolutional network (GCN) \cite{yuan2024node, wang2024gt} to obtain the spatial dependency. However, nodes and edges in dynamic graphs change both in the temporal and spatial dimensions \cite{b23, yuan2020multilayered}. This isolated modeling mechanism separates the temporal dimension from the spatial dimension, thereby blocking the spatio-temporal information propagation \cite{b19, b20, wu2023robust}.

Concretely, the defect of existing DGNNs may lead to the following limitations: a) the isolated treatment of the temporal dimension can result in the attenuation or eradication of interaction information among nodes; b) the extraction of spatio structural information may ignore the influence of temporal evolution. The above limitations are magnified when addressing a DG with complex graph structures \cite{he2024structure} and time-dependent features, thereby significantly affecting the accuracy of representation learning. Therefore, designing a DGNN capable of joint spatio-temporal information propagation is crucial for dynamic graph representation learning.

To address the above limitations, we design a novel information propagation method for joint spatio-temporal \cite{bi2024scg} information propagation based on the tensor M-product framework \cite{b24, b25}. Based on this, the tensorized lightweight graph convolutional network (TLGCN) is proposed to reduce memory consumption by omitting feature transformation and nonlinear activation. Our contributions are as follows:
\begin{itemize}
\item We design a novel information propagation method for joint propagation of spatio-temporal information based on the tensor M-product framework; 
\item We propose a tensorized lightweight graph convolutional network based on the above method, which significantly reduces the memory occupation of the model by omitting complex feature transformation and nonlinear activation.
\item We conduct comprehensive experiments on four widely used real-world dynamic graph datasets. Experimental results demonstrate that the proposed TLGCN consistently and significantly outperforms state-of-the-art models.
\end{itemize}

\begin{figure}[htbp]
\centerline{\includegraphics[width=0.5\columnwidth]{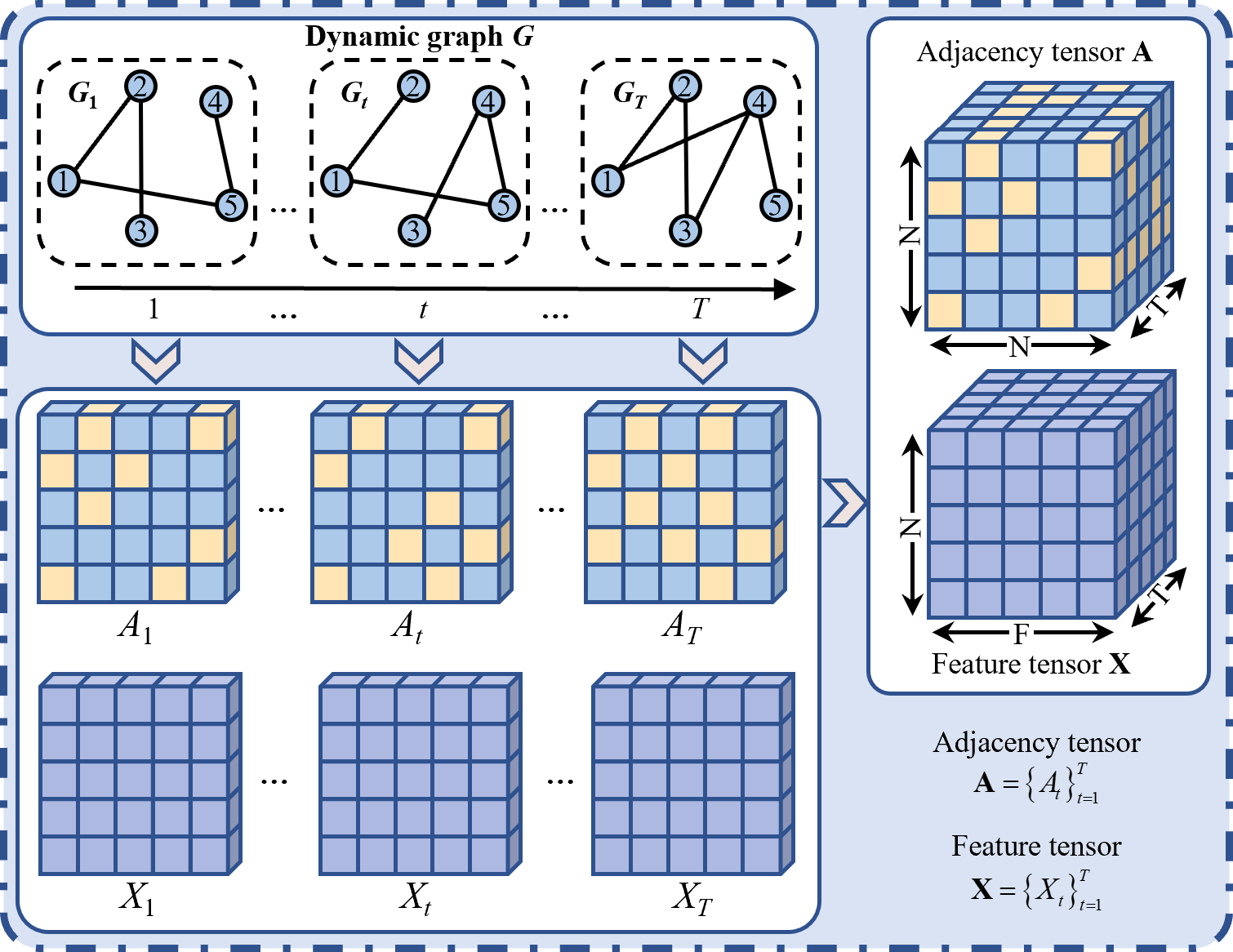}}
\caption{Third-order tensor form of a dynamic graph.}
\label{fig.1}
\end{figure}

\section{Related Work}
\subsection{Static Graph Neural Networks}
Static graph neural networks (SGNNs) have demonstrated powerful capabilities in modeling graph-structured data. For instance, GCN \cite{b8} leverages spatial-domain convolution to operate directly on graph structures, learning smooth node representations by aggregating neighborhood features. GraphSAGE \cite{b28} extends GCN by introducing an inductive learning framework that samples and aggregates features from local neighborhoods, making it scalable to large graphs and unseen data. GAT \cite{b29} employs an attention mechanism \cite{li2023saliency} to assign learnable weights to neighboring nodes, enabling adaptive feature aggregation based on node importance. DGI \cite{b30} optimizes node embeddings by maximizing the mutual information between local node features and global graph summaries, encouraging robust and informative representations. LightGCN  \cite{b31} simplifies the traditional GCN by removing feature transformation and nonlinear activation operations but retains the core component of neighbor aggregation, which makes it more suitable for recommendation tasks \cite{luo2021fast, luo2021fast, shang2021alpha}. However, all the above models are designed for static graphs \cite{wu2023graph}, if the graph structure and node features remain unchanged during training and inference, thus making it difficult to effectively address the evolutionary characteristics of DGs.

\subsection{Dynamic  Graph Neural Networks}
Dynamic graph neural networks (DGNNs) have emerged as a potent instrument for learning dynamic graphs. For instance, WinGNN  \cite{b20} employs a randomized gradient aggregation approach over sliding windows to model the temporal patterns \cite{luo2021adjusting} of a DG. EvolveGCN  \cite{b22} adopts RNNs to update the weight parameters of GCNs across different graph snapshots, enabling dynamic adaptation and enhancing model performance. SEIGN  \cite{b32} constructs a three-component architecture consisting of GCN-like message passing, temporal GRU parameter adjustments, and a self-attention mechanism inspired by the transformer. DySAT [19] presents a stacked GAT architecture for spatial attention and Transformers for temporal attention. TFE-GNN  \cite{b33} utilizes multilayer GNNs and cross-gated to build a temporal fusion encoder. EAGLE \cite{b35} models complex coupled environments and exploits spatial-temporal patterns for out-of-distribution prediction. However, the independent treatment of the temporal dimension in the above models may lead to the attenuation or loss of interaction information between nodes, while the effect of temporal evolution may be ignored when extracting spatial structure information.

\section{Preliminaries}
\subsection{Problem Formulation}
A DG is characterized as a sequence of static graphs evolving over a series of time slots, which can be denoted as $ G={\{G_t\}}^T_{t=1} $. 
For a specific time slot t, $ G_t=(V, E_t, X_t) $ reflects the snapshot of $G$, where $V$ denotes the set of nodes and $ E_t$ represents the connectivity relationship between nodes. 
Taking into account the snapshot $G_t$, its adjacency matrix and node feature matrix are denoted as $ A_t \in \mathbb{R} ^{N\times N} $ and $ X_t \in \mathbb{R} ^{N\times F} $ , where \textit{N} is the number of nodes and \textit{F} is the dimension of node feature. 
Especially, the individual element $a_{ijt} \in \{ 0,1 \}$ of $A_t$ indicates the absence or presence of an edge between node \textit{i} and node \textit{j} at time slot \textit{t}.
Apparently, a DG can be represented as a third-order tensor with dimensions \textbf{‘node-node-time’}. 
Hence, an adjacency tensor 
$\textbf{A} = \{A_t\}^T_{t=1} \in \mathbb{R}^{N \times N \times T}$ 
and a node feature tensor
$\textbf{X} = \{X_t\}^T_{t=1} \in \mathbb{R}^{N \times F \times T} $
can be utilized to comprehensively model a DG, as illustrated in Fig. \ref{fig.1}.

\begin{figure*}[htbp]
\centerline{\includegraphics[width=0.95\textwidth]{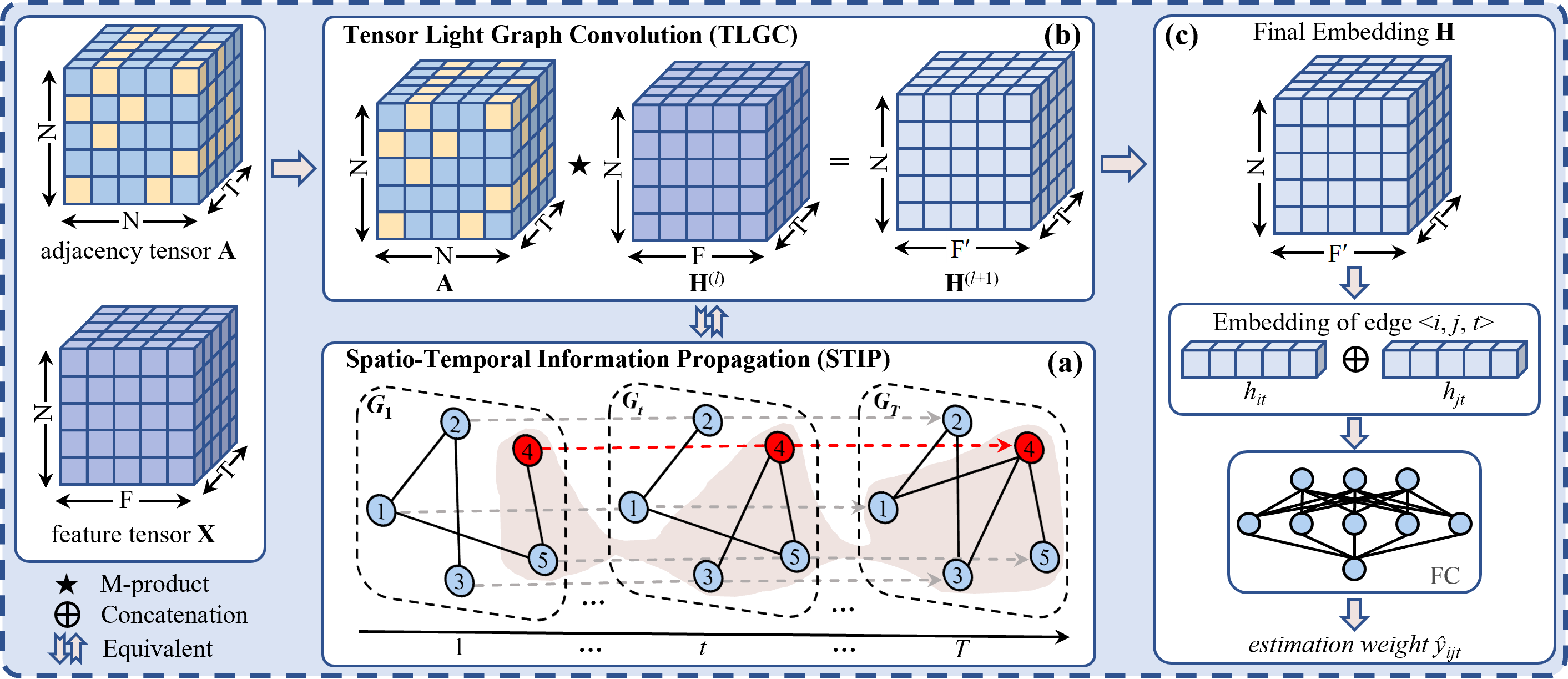}}
\caption{The overall framework of the proposed TLGCN.}
\label{fig.2}
\end{figure*}

In DGs, the weight estimation task aims to generate a predicted weight value using the adjacency tensor and the feature tensor. Specifically, the adjacency tensor and feature tensor are first processed by an encoder \cite{wu2024outlier, bi2023fast} to obtain node embedding; subsequently, the embedding is input into the estimation function to obtain the final estimation weight value. For the weight estimation task at time slot \textit{t}, the predicted weight value between the node pairs $<i,j>$ is calculated as follows:
\begin{align}
\left\{
    \begin{aligned}
	\textbf{H}&= \varepsilon(\textbf{A}, \textbf{X})\\
	\hat{y}_{ijt}&=f(h_{it},h_{jt})
    \end{aligned}
\right. ,
\end{align}
where $\varepsilon$ denotes the encoder, \textbf{H} denotes the node embedding tensor, $\textit{h}_{it}$ and $\textit{h}_{jt}$ denote the node embedding vector of node \textit{i} and node \textit{j} at time slot \textit{t} of \textbf{H} , respectively, 
$f(\cdot)$ denotes an estimation function, and $\hat{y}_{ijt}$ denotes the estimation weight value between node \textit{i} and node \textit{j} at time slot \textit{t}

\subsection{Tensor M-Product Framework}
This framework relies on a tensor operation known as the M-product \cite{b24, b25}, which defines the product of two three-dimensional tensors \cite{chen2024latent} resulting in another three-dimensional tensor. \textbf{Definitions 1-3} outline the specifics of the M-product.

\textbf{Definition 1: (Tensor M-transform)} Let $M \in \mathbb{R}^{T \times T}$ be a matrix. The M-transform of a tensor $\textbf{X} \in \mathbb{R}^{N \times F  \times T} $ is denoted by $(\textbf{X} \times_3 M) \in \mathbb{R}^{N \times F  \times T}$ and defined element-wise as:
\begin{equation}
    (\textbf{X} \times_3 M)_{ijt} = \sum_{k=1}^T M_{tk}x_{ijk}.
\end{equation}

\textbf{Definition 2: (Tensor Face-wise Product)} Let $\textbf{X} \in \mathbb{R}^{N \times F \times T}$ and $\textbf{Y} \in \mathbb{R}^{F \times K \times T}$ be two tensors. The face-wise product is denoted by $(\textbf{X} \Delta \textbf{Y}) \in \mathbb{R}^{N \times K  \times T}$ and is defined face-wise as:
\begin{equation}
    (\textbf{X} \Delta \textbf{Y})_{t} = X_t Y_t.
\end{equation}

\textbf{Definition 3: (Tensor M-product)} Let $\textbf{X} \in \mathbb{R}^{N \times F \times T}$ and $\textbf{Y} \in \mathbb{R}^{F \times K \times T}$ be two tensors, and let $M \in \mathbb{R}^{T \times T}$ be an invertible matrix. The M-product is denoted by $(\textbf{X} \bigstar \textbf{Y}) \in \mathbb{R}^{N \times K  \times T}$ and is defined face-wise as:
\begin{equation}
    \textbf{X} \bigstar \textbf{Y} = ((\textbf{X} \times_3 M) \Delta (\textbf{Y} \times_3 M)) \times_3 M^{-1}.
\end{equation}

\subsection{Graph Convolutional Network}
According to previous research \cite{b8, b36, b37}, GCN consists of multiple layers of graph convolution, where the core of each layer of graph convolution \cite{bi2024graph, bi2024graph} lies in the information propagation mechanism. Specifically, the information propagation mechanism \cite{bi2022two} captures features in the graph structure by aggregating the neighborhood information of the nodes and gradually updating the representation of each node. For a snapshot $G_t$ of a DG, the expression for a GCN can be described as:
\begin{equation}
{H_t}^{(l + 1)} = \sigma \left( {{{\tilde A}_t}{H_t}^{(l)}{W_t}^{(l)}} \right),
\end{equation}
where $\tilde{A_t} = D^{-\frac{1}{2}}(A_t+I_t)^{-\frac{1}{2}}$ denotes the symmetric normalized adjacency matrix, $(A_t+I_t)$ denotes the adjacency matrix with added self-loops, \textit{D} is the degree matrix of $(A_t+I_t)$, and $\sigma(\cdot)$ denotes a nonlinear activation function, $H_t^{(l+1)}$, $H_t^{(l)}$, and $W_t^{(l)}$ are the output node embedding matrix, the input node embedding matrix and the trainable weight matrix of the \textit{l}-th layer, respectively. The initial embedding matrix is derived from the node features, i.e., $H_t^{(0)}=X_t$.

\begin{figure}[htbp]
\centerline{\includegraphics[width=0.45\columnwidth]{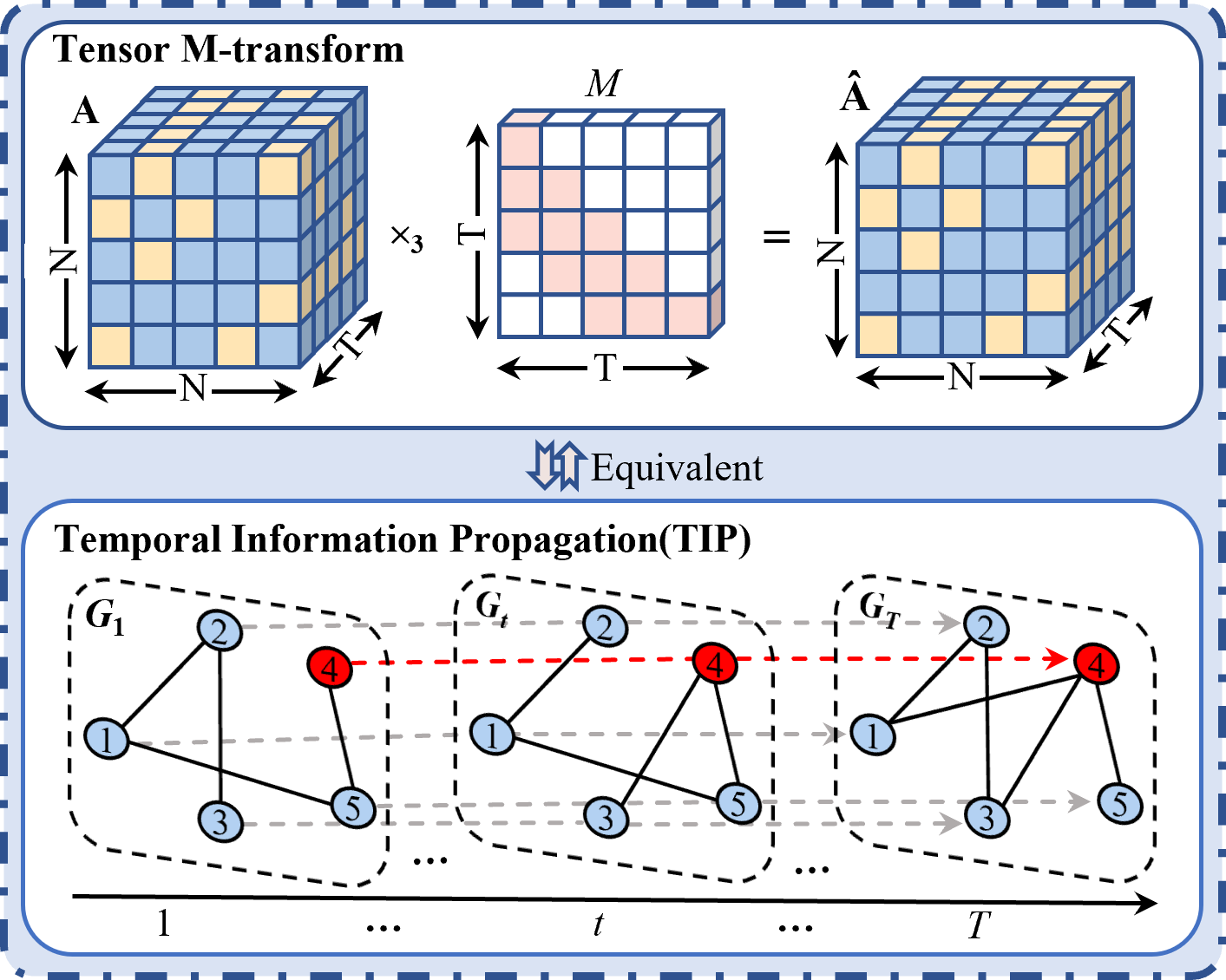}}
\caption{The temporal information propagation.}
\label{fig.3}
\end{figure}

\section{Methodology}
In this section, we present the proposed TLGCN model, which is illustrated in Fig. \ref{fig.2}. The framework is composed of two main components: (a) \textbf{\textit{Spatio-Temporal Information Propagation} (STIP) \textit{Module}}, which leverages tensor M-product framework for joint propagation of spatio-temporal information. (b) \textbf{\textit{Tensorized Lightweight Graph Convolution} (TLGC) \textit{Module}}, which is constructed based on the STIP module in combination with GCN, simplifies the convolution process by omitting the feature transformation and nonlinear activation of GCN. (c) \textbf{\textit{Weight Estimation module}}, which converts the final node embeddings into edge embeddings, then feeds edge embeddings into a fully connected layer to predict the weight values of the missing edges.

\subsection{Spatio-Temporal Information Propagation Module}
In DG representation learning, joint propagation of temporal and spatial information more effectively models DGs, capturing complex spatio-temporal interaction patterns. Existing DGCNs typically combine SGNN and SNN, where spatial information is propagated by SGNN and temporal information is handled by SNN. However, this approach fails to achieve effective temporal information propagation, leading to incomplete modeling of spatio-temporal interactions. To address this limitation, we design the STIP module based on the tensor M-product framework for joint spatio-temporal information propagation.

\textbf{Temporal information propagation:} In the STIP, the propagation of temporal information is 
realized through the M-transform operation, as shown in Fig. \ref{fig.3}. This operation aggregates the information of the adjacency matrices of different snapshots and passes it to the current snapshot, thus effectively capturing the temporal dependence in the DG. Specifically, the M-transform operation applies to the temporal dimension of the input tensor, i.e., propagates information along the temporal dimension. Mathematically, the temporal information propagation of the adjacency tensor \textbf{A} can be defined as:
\begin{equation}
{\bf{\hat A}} = {\bf{A}}{ \times _3}M ,
\end{equation}
where $\hat{\textbf{A}}$ and $\textbf{A}$ denote the adjacency tensor after applying the M-transform and the original adjacency tensor, respectively, and \textit{M} denotes the transformation matrix. For a clearer understanding, (6) can be expanded into vector form as follows for node \textit{i} at time slot \textit{t}: 
\begin{equation}
{\hat a_{it}} = \sum\limits_{k = 1}^T {{m_{tk}}{a_{ik}}} ,
\end{equation}
where $\hat a_{it}$ represents the adjacency vector of node \textit{i} at time slot \textit{t} after temporal information propagation and $m_{tk}$ represents the element from the transformation matrix \textit{M}. From (7), it can be seen that the tensor $\hat{\textbf{A}}$ realizes the effective propagation of temporal information by aggregating the information through the M-transform in the temporal dimension. Similarly, for the node embedding tensor $ \textbf{H}^{(l)}$  of the \textit{l}-th layer, the propagation 
of temporal information can also be realized by the M-transform, and the node embedding tensor after temporal information propagation is denoted as $ \hat{\textbf{H}}^{(l)}$.

\textbf{Spatial information propagation:} after completing the temporal information propagation, the spatial information propagation stage is entered, as shown in Fig. \ref{fig.4}. At this stage, the spatial dependence between nodes within the same snapshot is modeled by the tensor face-wise product between the adjacency tensor $\hat{\textbf{A}}$ and the node embedding tensor $ \hat{\textbf{H}}^{(l)}$ after the propagation of temporal information. It is calculated by the formula:
\begin{figure}[htbp]
\centerline{\includegraphics[width=0.45\columnwidth]{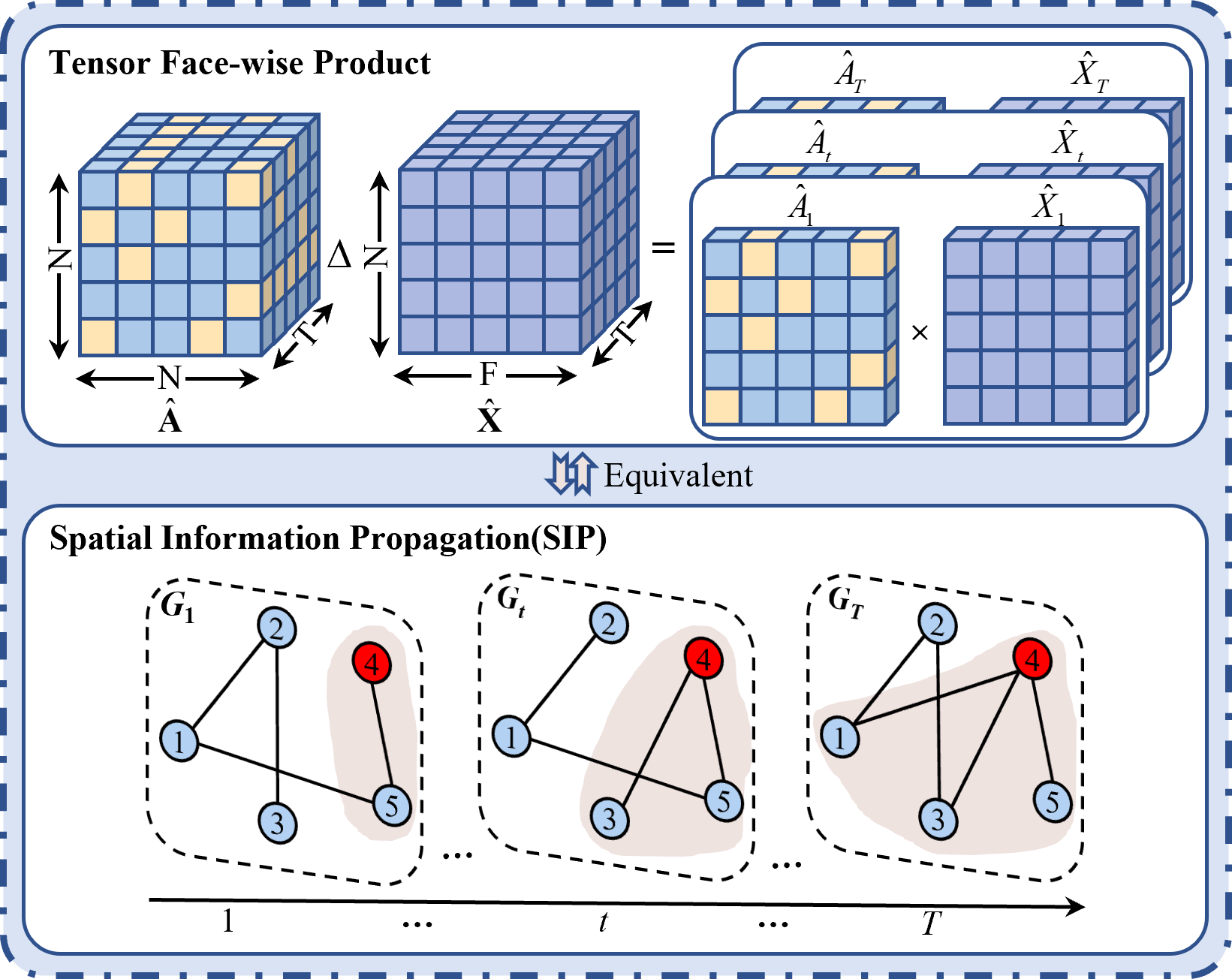}}
\caption{The spatial information propagation.}
\label{fig.4}
\end{figure}
\begin{equation}
{{\bf{H}}^{\left( {l + 1} \right)}} = {\bf{\hat A}}\Delta {{\bf{\hat H}}^{\left( l \right)}},
\end{equation}
where $\Delta$ represents the tensor face-wise product. This operation allows each node to aggregate information from its neighbors within the same snapshot, ensuring that local structural information is captured. In detail, for node \textit{i} at time slot \textit{t}, the vector representation is as follows:
\begin{equation}
h_{it}^{\left( {l + 1} \right)} = \sum\limits_{j \in \mathcal{N}\left( {i,t} \right)} {{{\hat a}_{ijt}}\hat h_{it}^{\left( l \right)}} ,
\end{equation}
where $\mathcal{N}\left( {i,t} \right)$ denotes all neighbors of node \textit{i} at time slot \textit{t} and node \textit{i} itself, $\hat{a}_{ijt}$ denotes the adjacency information between node \textit{i} and node \textit{j} at time slot \textit{t} after temporal information propagation, $ \hat{h}_{it}^{\left( l \right)} $ denotes the node embedding vector of node \textit{i} at time slot \textit{t} after temporal information propagation, and $ h_{it}^{\left( l+1 \right)} $ denotes the output node embedding vector of node \textit{i} in the \textit{l}-th layer spatial information propagation.

\subsection{Tensorized Lightweight Graph Convolution Module}
Given a DG, the input of TLGC is the node feature tensor \textbf{X} and the adjacency tensor \textbf{A}. In order to balance the information transfer between nodes and mitigate issues related to uneven degree distributions, we extend the symmetric normalization typically applied to adjacency matrices in conventional GCN to the adjacency tensor \textbf{A}. The symmetric normalization process is expressed mathematically as follows:
\begin{equation}
\tilde{\textbf{A}} = \textbf{D}^{-\frac{1}{2}} \Delta (\textbf{A} + \textbf{I}) \Delta \textbf{D}^{-\frac{1}{2}} ,
\end{equation}
where \textbf{I} denotes the identity tensor, \textbf{D} denotes the degree tensor of (\textbf{A}+\textbf{I}), $\Delta$ denotes the tensor face-wise product, $\tilde{\textbf{A}}$ denotes the symmetric normalization adjacency tensor. 

Then, according to the forward propagation process of a conventional GCN in (5) and the STIP module, the TLGC can be defined in the following form:
\begin{equation}
    \begin{split}
        \textbf{H}^{(l+1)} 
        &=\sigma(\tilde{\textbf{A}} \bigstar \textbf{H}^{(l)} \bigstar \textbf{W}^{(l)}) \\
        &= \sigma( ( \tilde{\textbf{A}} \times_3 M) \Delta (\textbf{H}^{(l)} \times_3 M) \Delta  (\textbf{W}^{(l)} \times_3 M) )
    \end{split} ,
\end{equation}
where $\textbf{H}^{(l+1)}$ and $\textbf{H}^{(l)}$ denote the output node embedding tensor and input node embedding tensor of the \textit{l}-th layer, respectively, the initial embedding tensor is derived from the node feature tensor, i.e., $\textbf{H}^{(0)} = \textbf{X}$, $\bigstar$ denotes the M-product, \textit{M} denotes the transformation matrix of the temporal information propagation. It should be noted that the M-transform is mainly responsible for the propagation of temporal information, so there is no need to undo it. Based on this, we remove the last inverse transformation $M^{-1}$. 

To reduce memory consumption and simplify implementation, we remove feature transformation and nonlinear activation, inspired by LightGCN \cite{b31}. TLGC’s simplified design focuses on directly aggregating neighborhood information, which reduces feature redundancy and preserves the dissimilarity of node features, thus mitigating excessive smoothing. The core idea is to allocate computational resources to the information propagation mechanism, maximizing model performance and efficiency. The final TLGC module is defined as follows:
\begin{equation}
    \textbf{H}^{(l+1)} = \tilde{\textbf{A}} \bigstar \textbf{H}^{(l)}= (\tilde{\textbf{A}} \times_3 M) \Delta (\textbf{H} \times_3 M).
\end{equation}

\textbf{Choice of the transformation matrix \textit{M}:} The transformation matrix \textit{M} controls the influence of each snapshot on the current snapshot, playing a key role in spatio-temporal information propagation. Therefore, setting \textit{M} appropriately is crucial for model performance. In this study, \textit{M} is set as a banded lower triangular matrix, where the lower triangular structure aggregates only historical snapshots to prevent future information leakage, and the banded structure limits the aggregation scope to improve computational efficiency. We consider two variants of the lower banded triangular \textit{M} matrix, with the first matrix $M_1$ defined as:
\begin{equation}
{m_{tk}} = \left\{ {\begin{array}{*{20}{c}}
{\frac{1}{{\min \left( {b,t} \right)}}}&{\max \left( {1,t - b + 1} \right) \le k \le t;}\\
0&{otherwise.}
\end{array}} \right.
\end{equation}
where \textit{b} denotes the  bandwidth of \textit{M}. This setting ensures that for any snapshot $G_t$, the elements of the \textit{t}-th row satisfy $\sum_k m_{tk} = 1$, i.e., all \textit{b} snapshots have equal weights. The equal weighting mechanism inherent in the matrix  $M_1$ ensures that each snapshot within the specified range has the same contribution, thus unifying the temporal impact and improving the temporal stability of the model.

The second matrix $M_2$, is defined as follows:
\begin{equation}
{m_{tk}} = \left\{ {\begin{array}{*{20}{c}}
{\frac{1}{{t - k + 1}},}&{\max \left( {1,t - b + 1} \right) \le k \le t;}\\
{0,}&{otherwise.}
\end{array}} \right.
\end{equation}
In this setup, the weights of the \textit{b} snapshots gradually increase over time, with higher weights the closer to the current snapshot. This weight increment strategy reflects the time correlation and highlights the importance of recent data, thus enhancing the model’s sensitivity to dynamic changes.

\subsection{Weight Estimation Module}
The final embedding tensor $\textbf{H} \in \mathbb{R}^{N \times F^{\prime} \times T}$ of the TLGC output can now be used for various tasks such as weight estimation, link prediction, and node classification. In this study, we apply this embedding to the weight estimation task using (1). Given an edge between node \textit{i} and node \textit{j} at time slot \textit{t}, the estimation module is defined as:
\begin{equation}
{\hat y_{ijt}} = f({h_{it}}||{h_{jt}}),
\end{equation}
where $h_{it} \in \mathbb{R}^{F^{\prime}}$ and $h_{jt} \in \mathbb{R}^{F^{\prime}}$ represent the final node embedding vectors of node \textit{i} and node \textit{j} at time slot \textit{t}, respectively, the operator $\Vert$ represents the concatenation operation, $({h_{it}}||{h_{jt}}) \in \mathbb{R}^{2F^{\prime}}$ represents the edge embedding, and $f(\cdot)$ represents a fully connected layer that maps the edge embeddings to a one-dimensional output, thereby obtaining the final weight estimation weight value.

Finally, The weight estimation task is optimized by a smooth $L_1$ loss as follows:
\begin{equation}
{L_w} = \left\{ {\begin{array}{*{20}{c}}
{\frac{1}{{2\beta }}\sum\limits_{{y_{ijt}} \in \Lambda } {{{\left( {{y_{ijt}} - {{\hat y}_{ijt}}} \right)}^2},} }&{\left| {{y_{ijt}} - {{\hat y}_{ijt}}} \right| < \beta }\\
{\sum\limits_{{y_{ijt}} \in \Lambda } {\left( {\left| {{y_{ijt}} - {{\hat y}_{ijt}}} \right| - \frac{1}{2}\beta } \right)} ,}&{otherwise}
\end{array}} \right.,
\end{equation}
where $\hat{y}_{ijt}$ and  $y_{ijt}$ represent the estimation weight value and the ground truth value of node \textit{i} and node \textit{j} at time slot \textit{t}, $\Lambda$ denotes the training dataset. Additionally, $\beta$ specifies the threshold at which the loss function switches between $\beta$-scaled $L_1$ and $L_2$ loss. The value of $\beta$ must be positive, and in this study it takes the value of one.

It should be noted that most of the dynamic graph datasets lack node features. For this reason, the node feature tensor \textbf{X} is randomly initialized and optimized during the training process, while $L_2$ regularization \cite{li2022diversified, wu20211} is used to prevent overfitting:
\begin{equation}
{L_x} = \lambda \left\| {\bf{X}} \right\|_2^2 = \lambda \sum\limits_{i = 1}^N {\sum\limits_{f = 1}^F {\sum\limits_{t = 1}^T {x_{ift}^2} } } ,
\end{equation}
where $\lambda$ is the regularization coefficient and $x_{ift}$ is an element of the node \textit{i} at time slot \textit{t} in the feature \textit{f}. By combining (17) and (18), we obtain the overall objective function and adapt the Adam to train it in an end-to-end fashion:
\begin{equation}
L = {L_w} + {L_x} .
\end{equation}

\section{Experiments}
\subsection{Evaluation metrics}
To assess the effectiveness of the TLGCN model, we utilize estimation accuracy \cite{chen2022mnl, bi2023proximal, li2022momentum} as the evaluation metric. Typically, \textbf{mean absolute error (MAE)} \cite{b38, chen2024state, wu2022double} and \textbf{root mean squared error (RMSE)} \cite{li2023nonlinear, li2022diversified, chen2021hierarchical} are two commonly used metrics \cite{zhang2022error, li2025learning, luo2024pseudo} for gauging the estimation accuracy of a model, the formula of them as follows:
\begin{equation*}
    \begin{array}{c}
    MAE = {{\left( {\sum\limits_{{y_{ijt}} \in \Lambda } {\left| {{y_{ijt}} - {{\hat y}_{ijt}}} \right|} } \right)} \mathord{\left/
    {\vphantom {{\left( {\sum\limits_{{y_{ijt}} \in \Lambda } {\left| {{y_{ijt}} - {{\hat y}_{ijt}}} \right|}      } \right)} {card\left( \Omega  \right)}}} \right.
     \kern-\nulldelimiterspace} {card\left( \Omega  \right)}}, \\
     RMSE = \sqrt {{{\left( {\sum\limits_{{y_{ijt}} \in \Lambda } {{{\left( {{y_{ijt}} - {{\hat y}_{ijt}}} \right)}^2}} } \right)} \mathord{\left/
 {\vphantom {{\left( {\sum\limits_{{y_{ijt}} \in \Lambda } {{{\left( {{y_{ijt}} - {{\hat y}_{ijt}}} \right)}^2}} } \right)} {card\left( \Omega  \right)}}} \right.
 \kern-\nulldelimiterspace} {card\left( \Omega  \right)}}} ,
    \end{array}
\end{equation*}
\begin{table}[htbp]
    \caption{Experimental Dataset Details.}
    \centering
    \begin{tabular}{cccccc}
    \toprule
        Datasets & Edges & Nodes & Density & Time slots \\ \midrule
        Bitcoin-OTC & 35592 & 6005 & 0.000986 & 64 \\
        Bitcoin-Alpha & 24186 & 7604 & 0.000418 & 64 \\
        FB-Messages & 61734 & 1899 & 0.0171 & 80 \\
        Email & 332334 & 986 & 0.342 & 80  \\ \bottomrule
    \end{tabular}
\end{table}
where ${\hat y}_{ijt}$, and $y_{ijt}$ represent the weight estimation value and the ground truth value of node \textit{i} and node \textit{j} at time slot \textit{t}, respectively, $card(\cdot)$ represents the cardinality of an enclosed set, $| \cdot |$ calculates the absolute value of an enclosed number, and $\Omega$ denotes the validating dataset.

\subsection{Datasets}
The effectiveness of our model is demonstrated through experiments conducted on four real-world datasets.
\begin{itemize}
\item \textbf{Bitcoin-OTC} \cite{b42} is a who-trusts-whom network for people trading Bitcoin on the Bitcoin-OTC platform. In this network, edge weights indicate trust scores.
\item \textbf{Bitcoin-Alpha} \cite{b43} is like Bitcoin-OTC but users and trust ratings from a different trading platform.
\item \textbf{FB-Messages} \cite{b44} is a Facebook-like social network from an online student community at the University of California. In this network, edge weights indicate the total volume of communications.
\item \textbf{Email}\cite{b45} is based on email data from a large European research institution. In this network, edge weights indicate the total number of emails exchanged.
\end{itemize}

All datasets are summarized in TABLE I and split into ten equal subsets \cite{wu2017highly, yuan2018effects}. Eight subsets are used for training, one for validation, and one for testing, following an 80\%-10\%-10\% split. For datasets without node features, we initialize them using the Xavier Glorot method and treat them as learnable parameters during training.

\subsection{Baseline Models}
To validate the effectiveness of our proposed TLGCN model, we compare it with several baseline models in a weight estimation task. These baseline models include the static GCN \cite{b8}, GAT \cite{b29}, and the dynamic EvolveGCN \cite{b22}, DGCN \cite{b41}, and WinGNN \cite{b20}.

\subsection{Experimental Settings}
To ensure a fair comparison, all models involved in the study are configured with the same structures and parameter values as follows:
\begin{enumerate}[label=\alph*.]
    \item All models are deployed on a GPU-equipped computer featuring an NVIDIA GeForce RTX 3050 GPU card;
    \item The Adam optimizer is used to optimize all models;
    \item Training stops uniformly for all models after 300 iterations or 20 iterations without error improvement.
    \item The node feature tensor of each model is initialized by Xavier and set to learnable parameters with its node feature dimension fixed to \textit{F}=16;
    \item For all the compared models, a grid search is utilized for the learning rate $\eta$  = \{0.00005, 0.0001, 0.0005, 0.001, 0.005, 0.01, 0.05\} and $L_2$ regularization coefficient $\lambda$ = \{0.00001, 0.00005, 0.0001, 0.0005, 0.001, 0.005, 0.01, 0.05, 0.1, 0.5, 1\} to achieve the optimal results.
    \item Other hyperparameters are fine-tuned based on their original papers to achieve optimal performance across different datasets respectively.
\end{enumerate}

\subsection{Performance Comparison}
This subsection reports the results of the comparison between TLGCN and baseline models on the weight value estimation task. The results are presented in TABLE II. The following conclusions can be drawn:

\textbf{DGNNs outperform SGNNs on several metrics.} Experimental results show that dynamic graph models (such as EvolveGCN, DGCN, and WinGNN) mostly outperform static models (GCN and GAT) in terms of MAE \cite{luo2020position, li2021proportional, xin2019non} and RMSE \cite{song2022nonnegative} metrics on datasets such as Bitcoin-OTC and Email. For example, on the Bitcoin-OTC dataset, the MAE of DGCN is 1.631, which is lower than the 1.689 of GCN, and on the Email dataset, the RMSE of WinGNN is 3.689, which is better than the 3.756 of GCN, which indicates that the dynamic graph model has a significant advantage in capturing the temporal features, but the static model is still competitive in some scenarios.

\begin{table*}[htbp]
    \centering
    \caption{The comparison results on MAE and RMSE. The best and second-best results in each column are highlighted in \textbf{bold} font and \uline{underlined}, respectively. All experimental results are the average values obtained after five runs of the model.}
    \centering
    \begin{tabular}{ccccccccc}
    \toprule
          \textbf{Datasets} & \textbf{Metrics} & \textbf{GCN} & \textbf{GAT} & \textbf{EvolveGCN} & \textbf{DGCN} & \textbf{WinGNN} & \textbf{TLGCN-V1} & \textbf{TLGCN-V2} \\ \midrule
         
        \multirow{2}{*}{\textbf{Bitcoin-OTC}} & MAE & 1.689±0.024 & 1.735±0.030 & 1.720±0.019 & 1.631±0.006 & 1.769±0.012 & \textbf{1.486}±0.007 & \uline{1.498}±0.005 \\ 
        & RMSE & 3.320±0.033 & 3.263±0.005 & 3.249±0.009 & 3.163±0.006 & 3.464±0.013 & \uline{2.748}±0.007 & \textbf{2.746}±0.001 \\
        
        \multirow{2}{*}{\textbf{Bitcoin-Alpha}} & MAE & 1.432±0.002 & 1.471±0.015 & 1.431±0.009 & 1.418±0.007 & 1.422±0.011 & \uline{1.317}±0.007 & \textbf{1.316}±0.007 \\ 
        & RMSE & 2.829±0.001 & 2.742±0.006 & 2.774±0.040 & 2.694±0.012 & 2.772±0.015 & \textbf{2.441}±0.005 & \uline{2.449}±0.005 \\ 
        
        \multirow{2}{*}{\textbf{FB-messages}} & MAE & 1.096±0.004 & 1.183±0.023 & 1.075±0.005 & 1.121±0.003 & 1.087±0.001 & \textbf{1.064}±0.009 & \uline{1.074}±0.009 \\ 
        & RMSE & 2.953±0.012 & 2.938±0.017 & 2.892±0.010 & 2.904±0.002 & 2.921±0.004 & \uline{2.752}±0.012 & \textbf{2.752}±0.009 \\ 
        
        \multirow{2}{*}{\textbf{Email}} & MAE & 1.471±0.002 & 1.465±0.001 & 1.461±0.002 & 1.487±0.002 & 1.461±0.002 & \textbf{1.369}±0.004 & \uline{1.373}±0.006 \\ 
        & RMSE & 3.756±0.002 & 3.743±0.002 & 3.559±0.013 & 3.676±0.006 & 3.689±0.003 & \textbf{3.395}±0.009 & \uline{3.412}±0.009 \\  \bottomrule
    \end{tabular}
\end{table*}

\begin{figure}[htbp]
\centerline{\includegraphics[width=\columnwidth]{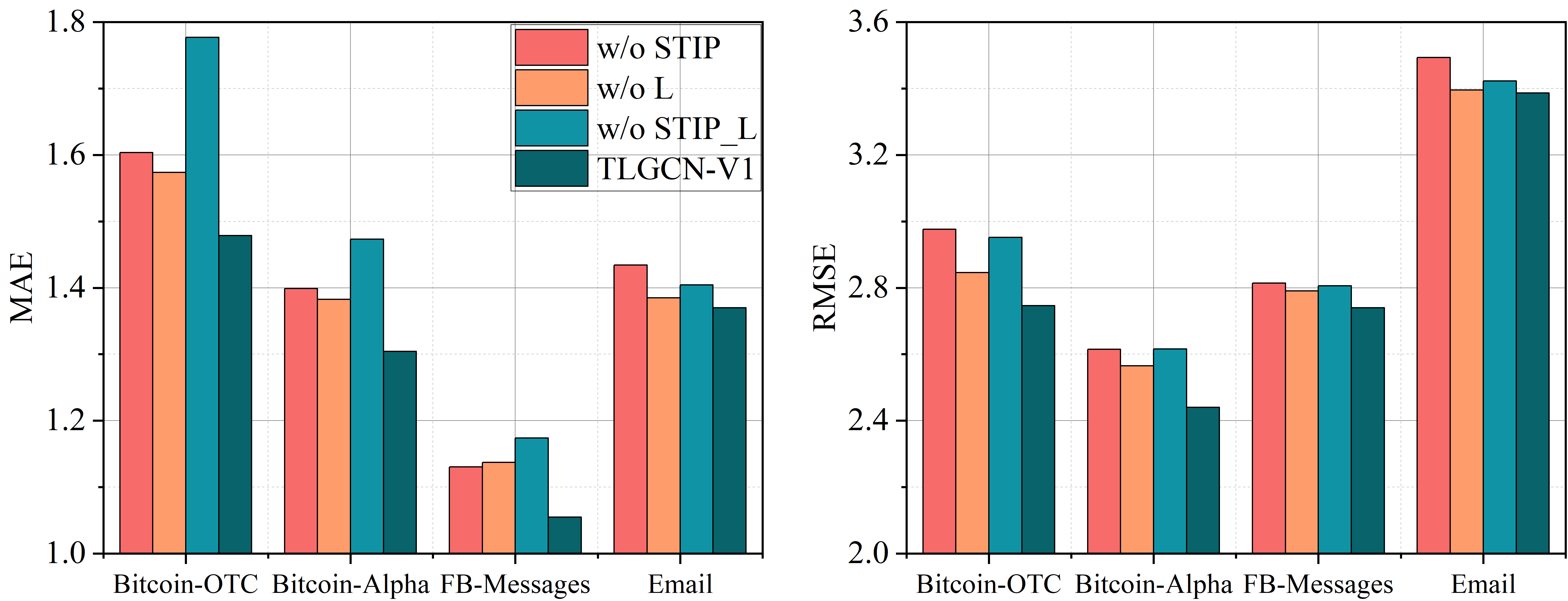}}
\caption{Ablation study.}
\label{fig.5}
\end{figure}

\begin{table}[htbp]
    \caption{The Memory Usage In MB.}
    \centering
    \begin{tabular}{cccccc}
    \toprule
     Datasets & Bitcoin-OTC & Bitcoin-Alpha & FB-Messages \\ \midrule
        w/o L & 487.6753 & 597.4541 & 202.6694 \\
        TLGCN-V1 & 326.1709 & 387.2715 & 137.5942 \\ \bottomrule
    \end{tabular}
\end{table}

\textbf{The two variants of TLGCN perform best among all compared models and lead on all datasets and metrics.} For example, on the FB-Messages dataset, the RMSE of both is 2.752, which records the optimal result; on the Bitcoin-Alpha dataset, the MAE of both TLGCN-V1 and TLGCN-V2 is 1.317, which outperforms all other models. This fully verifies the significant advantages of TLGCN in spatio-temporal information propagation and dynamic graph representation learning. It should be noted that while TLGCN-V2 enhances sensitivity to recent data through an incremental weighting strategy, the equal weighting mechanism of TLGCN-V1 provides a more stable and consistent performance on most datasets.

\subsection{Ablation study}

To investigate the impact of the STIP module and the Lightweight operation on the model, we perform an ablation analysis of TLGCN-V1. Specifically, we progressively remove both parts and create three TLGCN variants:
\begin{itemize}
    \item \textbf{w/o STIP}: In this variant, we remove the STIP module, adapt GCN as the base encoder, and maintain the lightweight operation of the feature transform and nonlinear activation.
    \item \textbf{w/o L}: In this variant, we remove the lightweight operation and maintain the STIP module.
    \item \textbf{w/o STIP\_L}: In this variant, we remove the STIP module and the lightweight operation and adapt GCN as the base encoder.
\end{itemize}

For the three variants, we keep all hyperparameters (e.g., learning rate $\eta$, $L_2$ regularization coefficient $\lambda$, dropout rate, etc.) the same as the optimal settings of TLGCN-V1. We also investigated the performance of TLGCN-V1 with the variant w / o L, which eliminates lightweight operation, in terms of maximum memory usage. The results, depicted in Fig. \ref{fig.5} and TABLE III, clearly show that:

\textbf{The STIP module plays a crucial role in capturing and propagating spatio-temporal dependencies in the model.} By enabling the spatio-temporal information propagation mechanism, STIP improves the model's ability to understand dynamic patterns and temporal relationships in the data. For example, on the Bitcoin-OTC dataset, the model with the STIP module demonstrates significantly lower MAE and RMSE compared to its variant without STIP, highlighting the importance of this module in improving the model's performance.

\textbf{The lightweight operation contributes to the model's accuracy by simplifying feature transformation and activation processes.} While its effect on MAE and RMSE may be less pronounced compared to the STIP module, its presence still enhances model performance. For instance, on the Bitcoin-Alpha dataset, the model with the lightweight operation yields slightly better MAE and RMSE results, indicating its role in improving the overall efficiency of the model. When both the STIP module and lightweight operation are combined, they work in tandem to significantly boost the model’s performance, as seen in the FB-Messages and Bitcoin-OTC datasets.

\begin{figure}[htbp]
\centerline{\includegraphics[width=0.7\columnwidth]{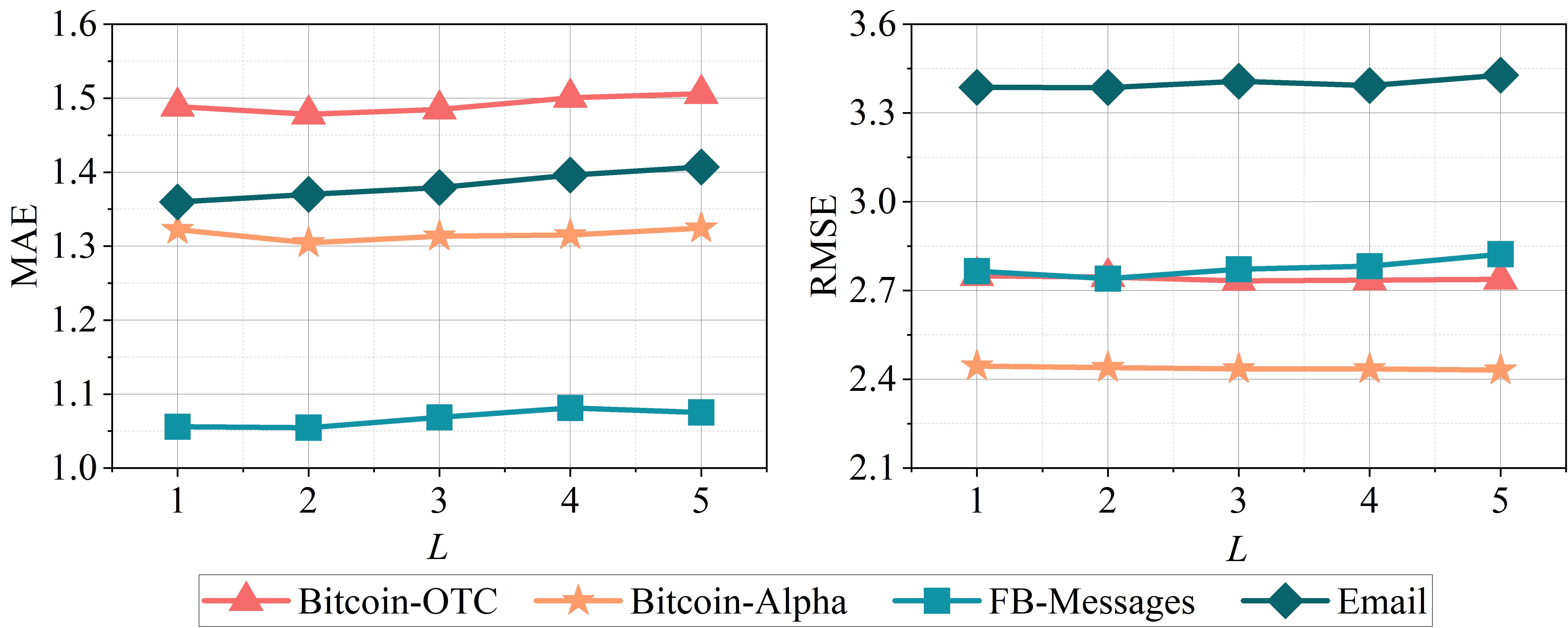}}
\caption{Hyper-parameter testing for different \textit{L}.}
\label{fig.6}
\end{figure}
\begin{figure}[htbp]
\centerline{\includegraphics[width=0.7\columnwidth]{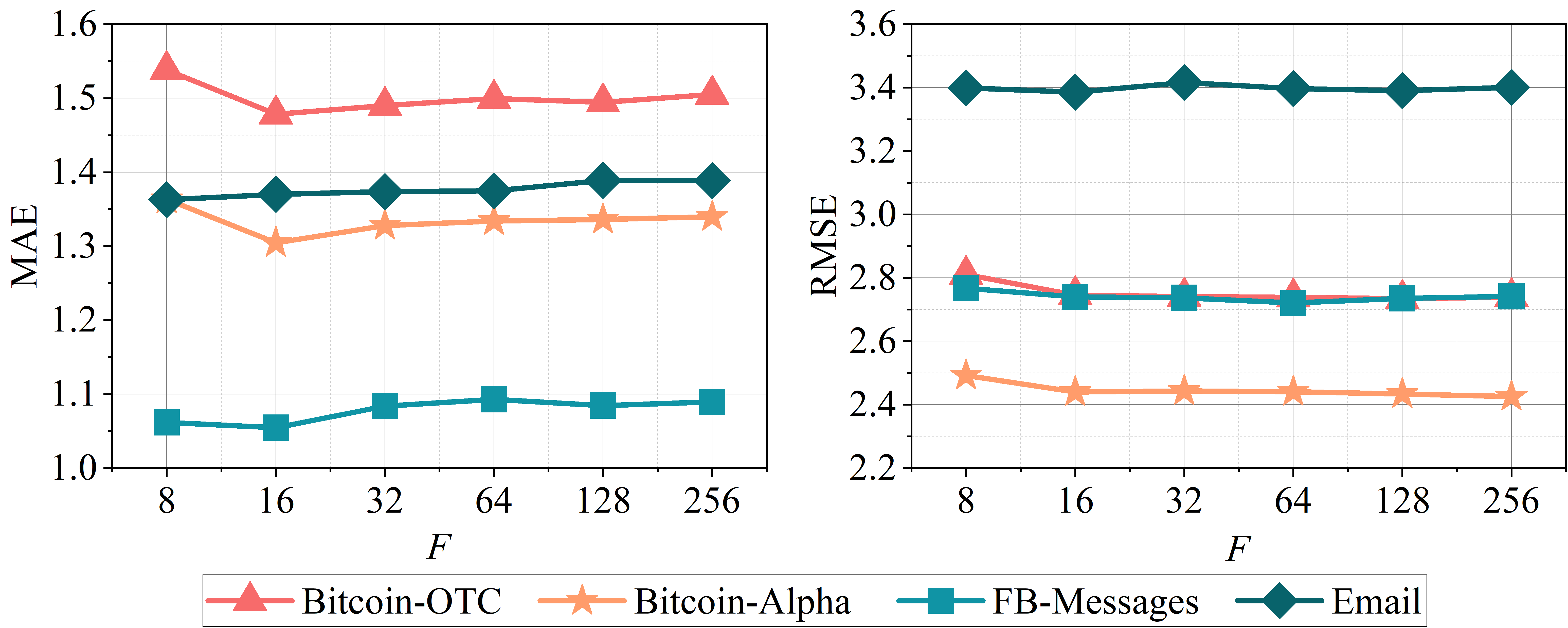}}
\caption{Hyper-parameter testing for different \textit{F}.}
\label{fig.7}
\end{figure}

\textbf{The lightweight operation significantly reduces the memory footprint compared to w/o \textit{L} variant on all datasets.} Specifically, it reduces the most on the Bitcoin-Alpha dataset by 35.19\%, reflecting its excellent memory optimization ability; it reduces the Email dataset by 28.74\%, showing significant optimization even with smaller data sizes. Overall, the lightweight operation is universal in terms of memory efficiency improvement, with an average reduction of about 30\%-35\%.
\subsection{Parameter Sensitivity Analysis} 
To thoroughly investigate the impact of different parameters in the TLGCN model, we conduct comprehensive parameter sensitivity experiments across all datasets. Specifically, we systematically explore the effects of varying the number of information propagation layers, denoted as \textit{L}, and the dimensions of node features, denoted as \textit{F}. For the number of information propagation layers, we consider a range of \{1, 2, 3, 4, 5\}, while for the dimensions of node features, the explored range includes \{8, 16, 32, 64, 128, 256\}. The detailed experimental results are summarized in Fig. \ref{fig.6} and Fig. \ref{fig.7}.

\textbf{Effect of Different $L$.} As the number of information propagation layers (\textit{L}) increases, model performance first improves and then declines. With $\textit{L}=1$, performance is limited by the small neighborhood information range. As \textit{L} increases to $\textit{L}=2$, the model captures both local and global features more effectively, achieving the best performance. Further increases in \textit{L} (e.g., $L \geq 3$) lead to a gradual decline in performance, though in some cases, performance may slightly improve or remain stable. This is due to excessive convolutional layers introducing noise, increasing the risk of over-smoothing, and complicating the model. Overall, $\textit{L}=2$ offers the best balance between accuracy and model complexity, capturing features effectively while avoiding unnecessary performance loss.

\textbf{Effect of Different $F$.} The experimental results show that the initialized feature dimension (\textit{F}) first improves and then levels off in terms of model performance. Increasing \textit{F} from a low value (e.g., $\textit{F}=8$) to a moderate value (e.g., $\textit{F}=16$) significantly boosts performance. However, further increases in \textit{F} (e.g., $F \geq 32$) result in diminishing returns, with slight performance degradation in some cases (e.g., FB-Messages dataset). This suggests that excessively high feature dimensions may introduce redundant information and increase computational overhead without significant performance gains. Therefore,$\textit{F}=16$ strikes the best balance between performance and computational efficiency.

\section{conclusions}
In this study, a novel tensorized lightweight graph convolutional network (TLGCN) is proposed, aiming to solve the problem of spatio-temporal information propagation isolation in traditional DGCNs. By designing a joint spatio-temporal information propagation method based on the tensor M-product framework, the effective combination of spatio-temporal dependencies is ensured. In addition, the proposed lightweight network significantly reduces the memory consumption of the model and improves the training efficiency by omitting the complex feature transformation and nonlinear activation. Experimental results show that TLGCN outperforms existing state-of-the-art models on four real-world dynamic graph datasets, validating its effectiveness and superiority in dynamic graph representation learning.


\bibliographystyle{ieeetr}     
\bibliography{TLGCN}     
\vspace{12pt}

\end{document}